\documentclass[10pt,twocolumn,letterpaper]{article}

\usepackage[pagenumbers]{cvpr}

\usepackage{cuted}

\usepackage{currfile}

\usepackage{caption}

\usepackage{multirow}
\usepackage{adjustbox}
\usepackage{colortbl}
\usepackage{microtype}
\usepackage{algorithm}
\usepackage{algpseudocode}

\renewcommand{\paragraph}[1]{\vspace{0.35em}\noindent\textbf{#1}\hspace{0.5em}}
\definecolor{oursbg}{RGB}{239,234,246}
\newcommand{\ours}{\rowcolor{oursbg}}
\usepackage{xspace}
\newcommand{\method}{FoldCache\xspace}
\newcommand{\methodlite}{FoldCache-Lite\xspace}
\newcommand{\dd}[1]{\,{\scriptsize\textcolor{gray}{(#1)}}}

\definecolor{cvprblue}{rgb}{0.21,0.49,0.74}
\usepackage[pagebackref,breaklinks,colorlinks,allcolors=cvprblue]{hyperref}

\def\confName{CVPR}
\def\confYear{2027}

\title{Fold the Loop: Decoding-Time KV Memory for Looped Vision-Language Models}
\author{Kun~Zhou\textsuperscript{1,2}\quad Wenmian~Yang\textsuperscript{3}\quad Zhensheng~Wang\textsuperscript{1,2}\quad Yanzeng~Li\textsuperscript{2}\quad Weijia~Jia\textsuperscript{2}\\[0.35em]
{\small \textsuperscript{1}School of Artificial Intelligence, Beijing Normal University, Beijing 100875, China}\\
{\small \textsuperscript{2}Institute of Artificial Intelligence and Future Networks, Beijing Normal University, Zhuhai 519087, China}\\
{\small \textsuperscript{3}School of Computer Science and Technology, Beijing Jiaotong University, Beijing 100044, China}}
\hypersetup{pdftitle={Fold the Loop: Decoding-Time KV Memory for Looped Vision-Language Models}, pdfauthor={Kun Zhou, Wenmian Yang, Zhensheng Wang, Yanzeng Li, Weijia Jia}}

\begin{document}
\maketitle

\begin{figure*}[t]
\centering
\includegraphics[width=\linewidth]{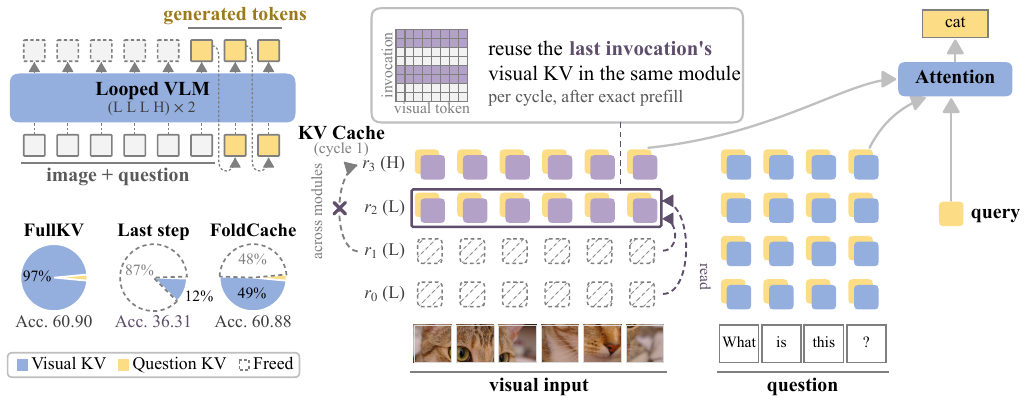}
\vspace{-1.4em}
\caption{
\textbf{Folding the visual KV of a looped VLM.}
LoopVL runs its two modules eight times, $(\mathrm{L\,L\,L\,H})\times 2$, and each invocation stores its own visual KV.
After an exact prefill, \method keeps only the visual KV of the last invocation of each module in each cycle.
The other invocations of that module read this copy (cycle~1 shown, where $r_0$ and $r_1$ read $r_2$) but never the copy of another module (crossed arrow), and question KV stays separate for every invocation.
Left: prefix KV memory on ChartQA and average accuracy over five benchmarks.
Last-step sharing from Ouro~\cite{ouro2025} serves every invocation from the final one, across modules and for question tokens, and loses 24.6 points, whereas \method stores 52\% of the FullKV prefix cache at the same accuracy.
}
\label{fig:teaser}
\end{figure*}

\begin{abstract}
Looped vision-language models (VLMs) deepen visual reasoning by repeatedly applying weight-tied modules. However, their decoding memory grows with recurrent depth and limits the image size and batch size that fit in a fixed memory budget. Each recurrent step produces its own key-value (KV) states, so the cache holds several copies of the same image. We find that, after an exact prefill, earlier steps still need access to visual information but can reuse the final visual KV within the same module. We propose FoldCache, which compresses visual KV along the recurrent axis while preserving the full prefill computation. FoldCache partitions recurrent steps within each module into groups and retains the final visual KV of each group for its members to share during decoding. Sharing stays within module boundaries, while question and answer KV remain separate for each step. The grouping granularity controls the degree of compression, and folding can be combined with token eviction and quantization to further reduce memory use. On LoopVL, FoldCache approximately halves prefix KV memory while matching full-cache accuracy on five open-ended benchmarks. These memory savings support batch sizes up to 1.7 times those of a full cache under the same GPU memory budget, and up to 2.6 times with the more aggressive FoldCache-Lite.
\end{abstract}

\section{Introduction}
\label{sec:intro}

Looped vision-language models deepen visual reasoning through additional computation in latent space.
They repeatedly apply weight-tied modules to update image and question representations without adding model parameters or generating longer reasoning traces~\cite{loopedlatent2025,huginn2025,ouro2025,loopvl2026,recursivevlm2026,loopvla2026,retrofitvlm2026}.
However, each recurrent invocation produces new key-value (KV) states that occupy cache memory during decoding.
For inputs with many visual tokens, increasing recurrent depth also means retaining more representations of the same image.
Higher image resolution makes each copy larger, so image size, recurrent depth and batch size compete for the same memory.
We ask whether deeper computation on an image requires keeping more visual states when generating the answer.

Existing methods reduce VLM inference costs through visual token pruning and merging~\cite{fastv2024,topv2025,visionzip2025,divprune2025,dycoke2025}, KV eviction, and quantization~\cite{vlcache2025,aircache2025,kivi2024}.
Cross-layer KV sharing further reduces state copies along depth~\cite{minicache2024,cla2024,yoco2024}.
Looped models add another dimension.
The same visual token passes through a shared module repeatedly and produces separate KV states at different recurrent steps (module invocations).
We refer to this dimension as the \emph{recurrent axis}.
Recent looped language models have explored sharing and compression along this axis~\cite{huginn2025,ouro2025,mor2025,lla2026,flashloop2026}, but report different conclusions about whether one step's KV can substitute for those of other steps~\cite{ouro2025,lla2026}.
In looped VLMs, long visual prefixes make such sharing more valuable but also harder.
Visual and textual states serve different roles, and independent modules form their own representations, so it is unclear which states can be shared and when.

We investigate this question by distinguishing how visual states are constructed from how they are used.
Prefill progressively builds the prefix representations of the image and question, whereas decoding reads these representations to generate an answer.
We find that decoding depends mostly on the visual KV of each module's final invocation, and that earlier invocations can read it in place of their own.
On ChartQA, masking the visual KV of a module's final invocation costs up to 34.0 points, and earlier invocations lose 1.7 points without visual KV but nothing when they read the final invocation's copy (\cref{fig:teaser}).
Earlier invocations thus still need the image, but they can read a copy that another invocation has already built.

This finding allows us to separate the computation required to construct representations from the storage required for decoding.
During prefill, the states produced at each recurrent step participate in subsequent updates.
Sharing KV prematurely changes this flow of information.
Keeping prefill exact ensures that the final visual states are still produced by the original recurrent computation.
Decoding can then read these completed states without retaining a separate visual copy for every invocation.

The reuse of visual states is further constrained by module boundaries and token types.
Recurrent invocations of the same module use the same KV projection weights, but shared weights alone do not make their states interchangeable.
Our analysis shows that extending visual KV sharing across independent modules harms accuracy.
Even within a module, also sharing question KV costs accuracy.
Visual sharing should occur after an exact prefill, respect module boundaries, and retain separate question and answer KV for each invocation.

We therefore propose FoldCache, which folds visual KV along the recurrent axis.
After the prefix has been constructed, FoldCache partitions the recurrent invocations of each module into groups.
Each group retains the visual KV from its final invocation for all members to read during decoding.
The model still executes the full recurrent computation, while the cache retains only the visual copies required by the grouping.
The grouping granularity determines the degree of compression.
By default, sharing occurs within each module and recurrent cycle, so a folded invocation stays close in depth to the copy it reads.
A more aggressive variant, FoldCache-Lite, shares across cycles and retains a single visual KV copy per module.

FoldCache complements cache compression methods that operate along other dimensions.
Token eviction reduces the sequence positions retained, quantization lowers the bit-width of each state, and folding reduces visual copies across recurrent invocations, so the three can be combined.
The saved space can accommodate more concurrent samples or inputs with more visual tokens without reducing the number of recurrent steps.

We evaluate FoldCache on the looped VLM LoopVL~\cite{loopvl2026}, covering open-ended question answering on charts, documents, and visual reasoning tasks, and verify that these tasks depend on visual KV during decoding.
A complementary study of text KV sharing in the looped language model Ouro~\cite{ouro2025} examines the sensitivity of different token types to sharing.
We further investigate aggressive folding, combinations with quantization and token eviction, and long-form answer generation.

Our main contributions are as follows.
\begin{itemize}
    \item We find that, in looped vision-language models, visual KV is redundant along the recurrent axis at decoding time.
    After an exact prefill, earlier recurrent steps can reuse the final visual KV of the same module, so their own visual KV need not be cached.

    \item We propose FoldCache, which compresses visual KV along the recurrent axis through grouped sharing within modules while preserving prefill and recurrent computation.
    The method supports adjustable compression granularity and can be combined with token eviction and quantization.

    \item Experiments on LoopVL show that FoldCache approximately halves prefix KV memory while matching full-cache accuracy on all evaluated open-ended benchmarks, and supports batch sizes up to $1.7\times$ those of a full cache under the same GPU memory budget.
\end{itemize}

\section{Related Work}
\label{sec:related}

\paragraph{Looped and recurrent-depth models.}
Weight sharing across depth dates back to the Universal Transformer and ALBERT, and is related to implicit-depth models and adaptive computation~\cite{universal2019,albert2020,deq2019,act2016,ponder2021}.
Looped Transformers can emulate programs, learn iterative algorithms, and reason in latent space, increasing reasoning depth without generating additional chain-of-thought tokens~\cite{giannou2023,yang2024looped,loopedlatent2025,snell2025,cot2022,deepseekr1_2025,coconut2024}.
Existing designs repeatedly invoke a single core block, adapt recursion depth per token, retrofit pretrained models, or alternate multiple weight-tied modules~\cite{huginn2025,ouro2025,mor2025,relaxedrecursive2025,hrm2025}.
Recurrence has also entered models built on standard vision-language pipelines~\cite{clip2021,llava2023,llava15_2024,qwen2vl2024,internvl2024}.
LoopVL alternates two modules, RecursiveVLM loops the entire language backbone, LoopVLA iteratively updates vision-language-action tokens, and other work constructs recurrent VLMs from pretrained backbones~\cite{loopvl2026,recursivevlm2026,loopvla2026,retrofitvlm2026}.
These works focus on modeling and training. We study the KV cost that recurrence adds at decoding time.

\paragraph{KV cache in looped models.}
Prior studies observe that recurrent steps need not each retain a separate cache.
Huginn shares a fixed budget of cache slots across iterations, MoR caches KV only at the first recursion, LPT lets later recurrences read memory written by the first, and retrofitted recurrent VLMs share a few cache copies across passes~\cite{huginn2025,mor2025,lpt2026,retrofitvlm2026}.
Ouro reports that KV reuse degrades accuracy during prefill but that last-step sharing is viable during decoding, without fully specifying which tokens are shared~\cite{ouro2025}.
LLA instead finds that last-loop reuse collapses generation and argues that the recurrent axis is low-rank but not directly collapsible~\cite{lla2026}.
Other methods compress this axis through a trained gated cache, hybrid latent attention, or quantization of state residuals between adjacent loops~\cite{melt2026,hla2026,flashloop2026}.
These studies consider a single looped stack, and all but one consider text models.
We examine how the sharing stage, module boundaries, and token types affect sharing, and derive a rule for folding visual KV within modules.
With a single module and module-wise grouping, this rule reduces to last-step sharing of visual KV.

\paragraph{Efficient VLM inference.}
Visual tokens typically occupy most of the VLM context.
Pruning and merging methods reduce redundant visual tokens before or inside the language model using attention, diversity, or similarity~\cite{fastv2024,sparsevlm2025,pyramiddrop2025,topv2025,divprune2025,tome2023,prumerge2024,visionzip2025}, with extensions to video and stage-aware compression~\cite{dycoke2025,sparsevila2025}.
Efficient encoders reduce tokens at the source, while multimodal KV compression evicts, merges, or selects cached entries through scoring~\cite{fastvlm2025,vlcache2025,aircache2025,lookm2024}.
These methods compress the token axis. FoldCache compresses the recurrent axis and can be combined with them.

\paragraph{KV cache compression.}
In language models, KV eviction retains heavy-hitter, attention-sink, or query-relevant entries, query-aware sparsity loads only relevant cache pages, and quantization reduces bit-width~\cite{h2o2023,streamingllm2024,scissorhands2023,fastgen2024,snapkv2024,pyramidkv2024,quest2024,kivi2024,kvquant2024}.
Architectural designs reduce KV through sharing across attention heads or compression into a latent space~\cite{mqa2019,gqa2023,deepseekv2_2024}.
Along depth, MiniCache merges the KV of adjacent layers, while CLA and YOCO introduce cross-layer sharing by design~\cite{minicache2024,cla2024,yoco2024}.
Unlike adjacent layers, invocations of the same module share weights, and we use module boundaries to define where sharing is allowed.
On the systems side, paged caching, IO-aware attention, and shared-prefix kernels improve cache efficiency~\cite{vllm2023,flashattention2022,flashattention2_2024,hydragen2024,flashinfer2025}.
Shared-prefix kernels reuse a prefix across requests, whereas we share visual KV across recurrent invocations within one request.

\section{Method}
\label{sec:method}

\begin{figure*}[t]
\centering
\includegraphics[width=\textwidth]{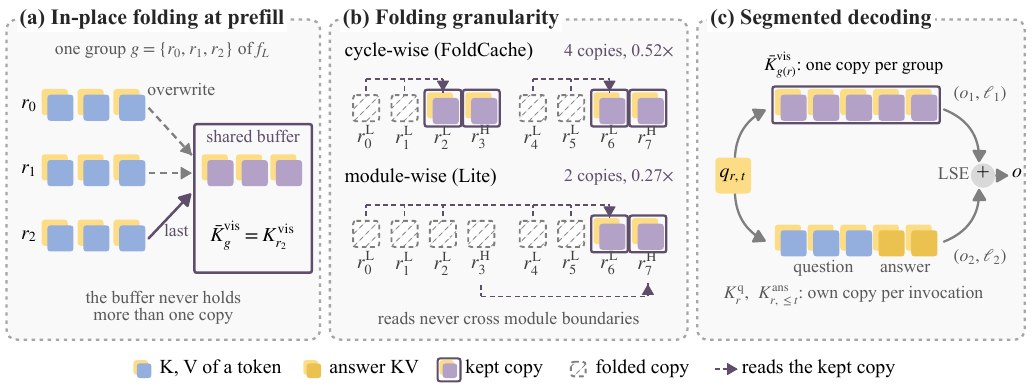}
\caption{\textbf{Overview of \method} on LoopVL.
(a) During prefill, the invocations of a group write their visual KV into one shared buffer in turn, so it ends with the last invocation.
(b) Cycle-wise folding (default) keeps $r_2,r_3,r_6,r_7$, and module-wise folding (\methodlite) keeps $r_6,r_7$. Groups never cross module boundaries.
(c) At decoding, each invocation of a sample attends to its group's visual segment, stored once per group and sample, and to its own question and answer KV, and the two results are merged by log-sum-exp.}
\label{fig:method}
\end{figure*}

\subsection{Preliminaries: Looped VLMs}
\label{sec:prelim}

A looped VLM concatenates $N_v$ visual tokens from a vision encoder, $N_q$ question tokens and the generated answer into one sequence, and processes it with $M$ weight-tied modules $f_1,\dots,f_M$, each with $L_s$ layers.
A schedule invokes these modules $R$ times in total. We index invocations by $r_0,\dots,r_{R-1}$ and write $m(r)$ for the module of invocation $r$ and $\mathrm{last}(m)$ for the last invocation of module $m$.
Single-stack looped LMs~\cite{huginn2025,ouro2025} have $M{=}1$, and hierarchical designs~\cite{hrm2025,loopvl2026} have $M{>}1$.
Only the output of the final invocation is passed to the language-model head.
The prefix (visual and question tokens) does not attend to the answer, so its states are fixed once prefill ends.

\paragraph{Instantiation: LoopVL.}
LoopVL~\cite{loopvl2026} uses $M{=}2$ modules, $f_L$ and $f_H$ with 16 layers each, and the H2L3 schedule
\begin{equation}
\underbrace{L\,L\,L\,H}_{\text{cycle 1}}\;\underbrace{L\,L\,L\,H}_{\text{cycle 2}},
\end{equation}
\ie $R=8$ invocations and 128 layer executions from 32 layers of parameters, with $\mathrm{last}(L)=r_6$ and $\mathrm{last}(H)=r_7$.
Its prefix uses bidirectional attention and answer tokens attend causally (PrefixLM).

\paragraph{KV cache of a looped model.}
At layer $l$ of invocation $r$, token $i$ produces
\begin{equation}
K_{r,l,i} = W_k^{m(r),l}\, h_{r,l,i},\qquad V_{r,l,i} = W_v^{m(r),l}\, h_{r,l,i}.
\end{equation}
The weights depend only on the module and layer, but the hidden state $h_{r,l,i}$ changes with $r$.
A KV cache therefore holds $R \cdot L_s$ per-layer entries, $R$ times more than a standard model with the same parameters.
Its prefix part is
\begin{equation}
\mathcal{M}_{\text{full}} = R \cdot L_s \cdot (N_v + N_q)\cdot 2d\cdot b,
\end{equation}
where $d$ is the hidden size and $b$ the bytes per element.
With $N_v\!\approx\!10^3$, visual KV accounts for over 90\% of the prefix cache (97\% on ChartQA, \cref{tab:main}).
Current looped VLMs avoid this cost by disabling the cache and recomputing all $R$ invocations over the full sequence for every generated token~\cite{loopvl2026}.
KV states thus multiply along the recurrent axis introduced in \cref{sec:intro}.

\subsection{Redundancy along the Recurrent Axis}
\label{sec:observations}

We probe the recurrent axis with invocation-level KV ablations on LoopVL (\cref{sec:analysis}) and summarize three findings.

\paragraph{(1) Prefill must stay exact, and decoding still reads the image.}
During prefill, visual and question tokens attend to each other in every invocation and are rewritten each time. Sharing visual KV across invocations at this stage breaks this interaction and costs 65.5 points on ChartQA.
After prefill, the prefix no longer changes and does not depend on the answer under the PrefixLM mask. Decoding still needs it, and evicting visual KV lowers ChartQA from 74.7 to 29.8.

\paragraph{(2) Redundancy follows the module boundary.}
Different modules have independent projections, so their KV states lie in different subspaces. In LoopVL, the cosine similarity between $f_L$ and $f_H$ keys is $\approx 0$.
Serving both modules from one shared copy costs 23--44 points on ChartQA (\cref{sec:ablation}).

\paragraph{(3) Decoding relies on the last invocation of each module.}
Masking the prefix in $r_6$ (last $L$) or $r_7$ (last $H$) costs 10.7 and 34.0 points on ChartQA. Earlier invocations still use the image, but reading the KV of the module's last invocation in place of their own leaves accuracy unchanged (\cref{sec:analysis}).

\paragraph{A first-order view.}
The three findings fit one account of the folding error.
When invocation $r$ reads a kept copy instead of its own visual KV, its attention output shifts by some $\delta_r$, and to first order the model output changes by $\sum_r \gamma_r\delta_r$, where $\gamma_r$ measures how much later computation depends on $r$.
The last invocations have the largest $\gamma_r$ (finding 3), so they keep their own copy.
The shift $\delta_r$ grows with the distance to the copy that $r$ reads. It is largest across modules, whose keys are nearly orthogonal (finding 2), and smaller within a cycle than across cycles (\cref{sec:ablation}).
In prefill, the shift is written back into the prefix and compounds over later invocations (finding 1).
It is also consistent with the conflicting reports on last-step reuse~\cite{ouro2025,lla2026}, which fails when a distant invocation or another module reads the copy.

\subsection{\method: Folding the Recurrent Axis}
\label{sec:foldcache}

\begin{algorithm}[t]
\caption{\method, per layer (layer index omitted). \textsc{Attn} returns the attention output and its log-sum-exp.}
\label{alg:foldcache}
\begin{algorithmic}[1]
\Require invocations $r_0,\dots,r_{R-1}$ and their groups $g(r)$
\Statex \textbf{Prefill}
\For{$r=r_0,\dots,r_{R-1}$}
  \State Run $r$ on the prefix as in the original model
  \State $\bar K^{\text{vis}}_{g(r)},\bar V^{\text{vis}}_{g(r)} \gets K^{\text{vis}}_{r},V^{\text{vis}}_{r}$ \Comment{in place}
  \State Store $K^{\text{q}}_{r},V^{\text{q}}_{r}$
\EndFor
\Statex \textbf{Decoding} (each new token)
\For{$r=r_0,\dots,r_{R-1}$}
  \State Append the new key and value to $K^{\text{ans}}_{r},V^{\text{ans}}_{r}$
  \State $o_1,\ell_1 \gets \textsc{Attn}(q_r,\bar K^{\text{vis}}_{g(r)},\bar V^{\text{vis}}_{g(r)})$
  \State $o_2,\ell_2 \gets \textsc{Attn}(q_r,[K^{\text{q}}_{r};K^{\text{ans}}_{r}],[V^{\text{q}}_{r};V^{\text{ans}}_{r}])$
  \State $o_r \gets (e^{\ell_1}o_1+e^{\ell_2}o_2)/(e^{\ell_1}+e^{\ell_2})$ \Comment{exact merge}
\EndFor
\end{algorithmic}
\end{algorithm}

Following these findings, \method keeps prefill unchanged and, after prefill, folds the visual KV along the recurrent axis with a tunable granularity (\cref{fig:method}).

\paragraph{Folding groups.}
We partition the invocations into groups $\mathcal{G}$ that never cross a module boundary, and let $\mathrm{last}(g)$ be the last invocation of group $g$.
For every group and layer, \method keeps only the visual KV of that last invocation,
\begin{equation}
\bar K^{\text{vis}}_{g,l} = K^{\text{vis}}_{\mathrm{last}(g),\,l},\qquad
\bar V^{\text{vis}}_{g,l} = V^{\text{vis}}_{\mathrm{last}(g),\,l},
\end{equation}
and every invocation $r\in g$ reads this copy.
Question-token KV and answer-token KV are kept exact for every invocation.
At decoding step $t$, invocation $r$ in group $g(r)$ attends to
\begin{equation}
\big[\,\bar K^{\text{vis}}_{g(r),l}\,;\,K^{\text{q}}_{r,l}\,;\,K^{\text{ans}}_{r,l,\le t}\,\big]
\end{equation}
and the corresponding values, so the visual cache holds $|\mathcal{G}|$ copies instead of $R$.

\paragraph{Two granularities.}
\emph{Cycle-wise} folding (\method, our default) groups the invocations of each module within each cycle.
For LoopVL this keeps $r_2,r_3,r_6,r_7$, \ie 4 visual copies, and stores about half of the prefix KV.
\emph{Module-wise} folding (\methodlite) groups all invocations of a module, keeping only $r_6$ and $r_7$ (2 copies, about a quarter of the prefix KV).
It saves more memory but is more sensitive on text-dense images (\cref{sec:main}).

\paragraph{Visual tokens only.}
Folding question tokens as well saves little memory but costs accuracy (\cref{sec:ablation}).

\paragraph{Implementation.}
\Cref{alg:foldcache} summarizes prefill and decoding.
The shared buffer of each group is overwritten in place during prefill, so it never holds more than one copy and ends with the last invocation.
At decoding, the shared visual base (one per module, or per module and cycle) is stored once and shared along the batch dimension. Each invocation attends to this base and to its own question and answer KV separately, and the two partial results are merged by their log-sum-exp (supplementary material).
The segments are never concatenated, so the shared copy is read in place.
\method needs no training and no change to the model, and it can be combined with token pruning and KV quantization.
Quantizing the retained copies reduces storage further. Realizing a speed-up from quantized KV would require dedicated kernels, which we do not implement.

\section{Experiments}
\label{sec:experiments}

\begin{table*}[t]
\centering
\small
\setlength{\tabcolsep}{6pt}
\caption{
\textbf{Main results} on LoopVL-1B.
\textbf{KV Mem.}: stored prefix KV relative to FullKV (ChartQA).
Gray: paired difference to FullKV of the same run.
DocVQA, InfoVQA and TextVQA use fixed 2000-sample subsets.
}
\label{tab:main}
\resizebox{\textwidth}{!}{\begin{tabular}{l c c c c c c c}
\toprule
\textbf{Method} & \textbf{KV Mem.}
& \textbf{ChartQA} & \textbf{MathVision} & \textbf{DocVQA} & \textbf{InfoVQA} & \textbf{TextVQA} & \textbf{Avg.} \\
\midrule
FullKV                                   & 1.00 & 74.72 & 38.26 & 78.44 & 60.28 & 52.81 & 60.90 \\
Visual KV eviction                       & 0.03 & 29.80\dd{$-$44.92} & 36.51\dd{$-$1.75} & 26.22\dd{$-$52.22} & 38.35\dd{$-$21.93} & 22.12\dd{$-$30.68} & 30.60\dd{$-$30.30} \\
\midrule
\multicolumn{8}{l}{\emph{Token-wise KV eviction}} \\
H2O~\cite{h2o2023}                       & 0.27 & 74.52\dd{$-$0.20} & 38.26\dd{0.00} & 77.78\dd{$-$0.66} & 60.00\dd{$-$0.28} & 52.77\dd{$-$0.04} & 60.67\dd{$-$0.23} \\
SnapKV~\cite{snapkv2024}                 & 0.27 & 74.68\dd{$-$0.04} & 38.19\dd{$-$0.07} & 78.02\dd{$-$0.42} & 59.92\dd{$-$0.36} & 52.02\dd{$-$0.79} & 60.57\dd{$-$0.33} \\
\midrule
\multicolumn{8}{l}{\emph{Cross-invocation KV cache sharing}} \\
Last-step KV sharing (Ouro)~\cite{ouro2025} & 0.13 & 48.40\dd{$-$26.32} & 31.18\dd{$-$7.07} & 32.92\dd{$-$45.52} & 41.30\dd{$-$18.98} & 27.73\dd{$-$25.08} & 36.31\dd{$-$24.59} \\
Zero-shot KV sharing, $k{=}2$ (Huginn)~\cite{huginn2025} & 0.25 & 70.24\dd{$-$4.48} & 36.64\dd{$-$1.61} & 61.44\dd{$-$17.00} & 54.81\dd{$-$5.46} & 42.48\dd{$-$10.32} & 53.13\dd{$-$7.78} \\
Zero-shot KV sharing, $k{=}4$ (Huginn)~\cite{huginn2025} & 0.50 & 71.12\dd{$-$3.60} & 36.91\dd{$-$1.35} & 66.31\dd{$-$12.13} & 56.69\dd{$-$3.59} & 43.70\dd{$-$9.11} & 54.95\dd{$-$5.95} \\
Adjacent-invocation merging (MiniCache)~\cite{minicache2024} & 0.55 & 73.96\dd{$-$0.76} & 38.22\dd{$-$0.03} & 72.16\dd{$-$6.28} & 57.38\dd{$-$2.89} & 51.80\dd{$-$1.00} & 58.70\dd{$-$2.20} \\
\midrule
\multicolumn{8}{l}{\emph{KV quantization}} \\
KIVI (INT4)~\cite{kivi2024}              & 0.28 & 74.80\dd{$+$0.08} & 38.09\dd{$-$0.16} & 78.45\dd{$+$0.01} & 60.28\dd{0.00} & 52.45\dd{$-$0.36} & 60.81\dd{$-$0.09} \\
KIVI (INT2)~\cite{kivi2024}              & 0.25 & 74.40\dd{$-$0.32} & 38.19\dd{$-$0.07} & 78.09\dd{$-$0.35} & 60.08\dd{$-$0.20} & 52.99\dd{$+$0.18} & 60.75\dd{$-$0.15} \\
FlashLoop (INT4, decoding only)~\cite{flashloop2026} & 0.28 & 74.88\dd{$+$0.16} & 38.42\dd{$+$0.16} & 78.34\dd{$-$0.09} & 60.18\dd{$-$0.09} & 52.80\dd{$-$0.01} & 60.93\dd{$+$0.03} \\
FlashLoop (INT2, decoding only)~\cite{flashloop2026} & 0.25 & 74.44\dd{$-$0.28} & 38.29\dd{$+$0.03} & 77.82\dd{$-$0.62} & 59.46\dd{$-$0.82} & 52.52\dd{$-$0.29} & 60.51\dd{$-$0.40} \\
\midrule
\ours \textbf{\method (cycle-wise, default)} & 0.52 & 74.76\dd{$+$0.04} & 38.19\dd{$-$0.07} & 78.15\dd{$-$0.29} & 60.40\dd{$+$0.12} & 52.91\dd{$+$0.10} & 60.88\dd{$-$0.02} \\
\method + INT4                           & 0.17 & 74.76\dd{$+$0.04} & 38.13\dd{$-$0.13} & 78.07\dd{$-$0.37} & 60.22\dd{$-$0.05} & 52.68\dd{$-$0.13} & 60.77\dd{$-$0.13} \\
\methodlite (module-wise)                & 0.27 & 74.64\dd{$-$0.08} & 38.10\dd{$-$0.16} & 78.26\dd{$-$0.18} & 60.05\dd{$-$0.23} & 51.91\dd{$-$0.90} & 60.59\dd{$-$0.31} \\
\methodlite + INT4                       & 0.10 & 74.60\dd{$-$0.12} & 37.99\dd{$-$0.26} & 78.18\dd{$-$0.26} & 59.79\dd{$-$0.49} & 51.87\dd{$-$0.94} & 60.49\dd{$-$0.41} \\
\bottomrule
\end{tabular}}
\end{table*}

\subsection{Experimental Setup}
\label{sec:setup}

\paragraph{Model.}
We use LoopVL-1B~\cite{loopvl2026}, to our knowledge the only looped VLM with public weights, with its native H2L3 schedule ($R=8$ module invocations).
Unless noted otherwise, all experiments run on NVIDIA A800 GPUs in BF16 with greedy decoding. Two ablations and the Ouro study run on RTX 4090D GPUs and are compared with FullKV of the same run.

\paragraph{Benchmarks.}
KV compression applied after prefill cannot change an answer that is fully determined by the first generated token, since that token is produced by the uncompressed prefill.
We therefore report open-ended benchmarks with multi-token answers as the main evidence: ChartQA~\cite{chartqa}, MathVision~\cite{mathvision}, DocVQA~\cite{docvqa}, InfoVQA~\cite{infovqa} and TextVQA~\cite{textvqa}.
Multiple-choice benchmarks (MMStar~\cite{mmstar}, AI2D~\cite{ai2d}, VMCBench~\cite{vmcbench} and RealWorldQA~\cite{realworldqa}) are reported in the supplementary material as a sanity check. Neither \methodlite nor evicting all visual KV changes their accuracy.
Prompts and metrics follow VLMEvalKit~\cite{vlmevalkit}: relaxed accuracy for ChartQA, ANLS for DocVQA and InfoVQA, and VQA accuracy for TextVQA.
DocVQA, InfoVQA and TextVQA use the validation splits released by lmms-lab.
A benchmark is regarded as sensitive to KV compression only if visual KV eviction degrades it by at least 5 points.

\paragraph{Baselines.}
\emph{FullKV} caches the KV states of every invocation, and its outputs match those of LoopVL decoding without a cache.
\emph{Visual KV eviction} removes all visual KV after prefill and serves as a lower bound.
We adapt three families of methods to LoopVL.
Token-wise KV eviction: H2O~\cite{h2o2023} and SnapKV~\cite{snapkv2024}, applied to the visual KV of every invocation at the same memory budget as \methodlite. Tokens are selected independently in each invocation, using the last 8 question tokens as the observation window, and the prefix attends bidirectionally.
Cross-invocation KV cache sharing: last-step KV sharing from Ouro~\cite{ouro2025} and zero-shot KV sharing from Huginn~\cite{huginn2025}, both sharing the whole prefix as in their original form, and MiniCache~\cite{minicache2024} transferred from adjacent layers to adjacent invocations within a module (SLERP merging, 5\% of tokens kept unmerged).
KV quantization: per-invocation INT4 and INT2 quantization with KIVI~\cite{kivi2024} (INT2 in its original configuration: group size 32, full-precision residual of 128 tokens), and the decoding-stage residual quantization of FlashLoop~\cite{flashloop2026}.
All methods keep prefill uncompressed and compress only the prefix KV cache. Variants that also share answer-token KV are reported in the supplementary material.
\emph{Memory} is the stored prefix KV on ChartQA relative to FullKV (visual, header and question tokens, including quantization parameters).
\emph{Avg.} is the simple mean over the five benchmarks.

\paragraph{Statistical protocol.}
All methods are evaluated on the same samples within the same run.
We report paired differences to FullKV and test non-inferiority with a paired bootstrap and a $-1.5$-point margin on every sensitive benchmark.

\subsection{Main Results}
\label{sec:main}

\Cref{tab:main} compares \method with FullKV and the adapted baselines. All numbers are from A800 GPUs and paired against FullKV of the same run.
Visual KV eviction degrades ChartQA, DocVQA, InfoVQA and TextVQA by 44.9, 52.2, 21.9 and 30.7 points, confirming that these benchmarks are sensitive to compressing visual KV.

\paragraph{Default granularity.}
\method (cycle-wise) keeps 4 of the 8 visual KV copies, about half of the prefix KV, and matches FullKV on all five benchmarks, passing the non-inferiority test on each. All paired differences lie between $-0.29$ and $+0.12$, and its five-benchmark average is within 0.02 points of FullKV.
On the full DocVQA, InfoVQA and TextVQA validation sets, it also stays within 0.1 points of FullKV.
At similar memory, merging adjacent invocations within a module with MiniCache loses up to 6.3 points on DocVQA and 2.2 points on average.
Quantizing the retained copies to INT4 further reduces the prefix KV to 16--17\% (27\% on MathVision), below the 28\% of INT4 quantization alone, at comparable accuracy.
In long chain-of-thought generation (ChartQA, official reasoning trigger, up to 512 tokens), \method is non-inferior to FullKV (supplementary material).

\paragraph{Aggressive granularity.}
\methodlite keeps one copy per module and stores 27\% of the prefix KV.
It stays within 0.23 points of FullKV on four benchmarks and loses 0.90 points on text-dense TextVQA (full-set results in the supplementary material).
In long chain-of-thought generation, it falls just short of the non-inferiority margin and loses 1.60 points on the human-written ChartQA questions, in line with TextVQA.
We therefore use the cycle-wise granularity by default and \methodlite when memory is tight.

\paragraph{Cross-invocation sharing.}
Strategies transferred from looped LMs fall far behind on every sensitive benchmark. Last-step sharing (Ouro) loses 19--46 points and zero-shot sharing (Huginn) 3.6--17 points, with the largest drops on document and scene-text images.
\Cref{sec:ablation} attributes their losses to the two design rules of \method.

\begin{figure}[t]
\centering
\includegraphics[width=\linewidth]{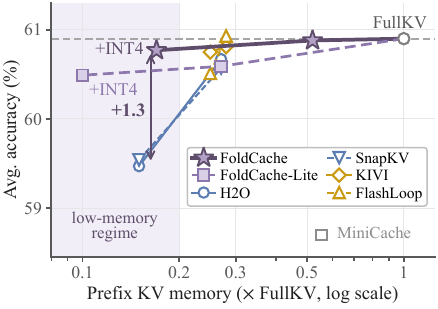}
\caption{\textbf{Accuracy--memory trade-off}, averaged over the five benchmarks of \cref{tab:main}.
Token-wise eviction is shown at 10\% and 25\% of the visual tokens and quantization at INT2 and INT4.}
\label{fig:pareto}
\end{figure}

\paragraph{Token-wise eviction and quantization.}
At similar memory, token-wise eviction and INT4 quantization are on par with \methodlite.
They compress the token and bit-width axes and can be combined with \method (\cref{fig:pareto}).

\paragraph{Accuracy--memory trade-off.}
Folding, token eviction and quantization compress different axes, so they stack.
At $0.25\times$ of the prefix KV or more, all compression methods stay close to FullKV and are largely on par.
The differences appear below $0.2\times$, where folding adds to quantization. \method{}\,+\,INT4 ($0.17\times$, 60.77) and \methodlite{}\,+\,INT4 ($0.10\times$, 60.49) stay within 0.5 points of FullKV, and token-wise eviction can be added on top (\cref{tab:compose}). Token-wise eviction alone at a similar budget ($0.15\times$, 10\% of the visual tokens) is 1.0--1.3 points lower (H2O 59.47, SnapKV 59.54).
\method{}\,+\,INT4 also stores less than INT2 quantization in KIVI's recommended configuration (about $0.16\times$ versus $0.24\times$ on DocVQA, InfoVQA and TextVQA) and matches or exceeds their accuracy in all 10 paired comparisons with KIVI-INT2 and FlashLoop-INT2 (supplementary material).

\begin{table}[t]
\centering
\footnotesize
\setlength{\tabcolsep}{4pt}
\caption{\textbf{Composing the three axes} with \methodlite (RTX 4090D).
Accuracy change relative to FullKV of the same run; SnapKV keeps 25\% of the visual tokens.
\textbf{Mem.} is computed analytically and includes token indices.}
\label{tab:compose}
\begin{tabular}{c c c c c c}
\toprule
\textbf{Fold} & \textbf{SnapKV} & \textbf{INT4} & \textbf{Mem.} & \textbf{ChartQA} & \textbf{InfoVQA} \\
\midrule
$\checkmark$ &              &              & 0.27  & $-$0.20 & $-$0.34 \\
             & $\checkmark$ &              & 0.27  & $-$0.48 & $-$0.13 \\
\midrule
$\checkmark$ &              & $\checkmark$ & 0.09  & $-$0.20 & $-$0.60 \\
             & $\checkmark$ & $\checkmark$ & 0.09  & $-$0.52 & $-$0.24 \\
$\checkmark$ & $\checkmark$ &              & 0.15  & $-$0.72 & $-$0.30 \\
\midrule
\ours $\checkmark$ & $\checkmark$ & $\checkmark$ & 0.06 & $-$0.64 & $-$0.43 \\
\bottomrule
\end{tabular}
\end{table}

\paragraph{Composing the three axes.}
Adding token-wise eviction as a third axis (\cref{tab:compose}), \methodlite with SnapKV and INT4 keeps 6\% of the prefix KV and stays within the non-inferiority margin on ChartQA ($-0.64$) and InfoVQA ($-0.43$).
Since only visual KV is folded, the saving grows with image resolution (\cref{fig:efficiency}).

\subsection{Invocation-Level Analysis}
\label{sec:analysis}

\begin{figure}[t]
\centering
\includegraphics[width=\linewidth]{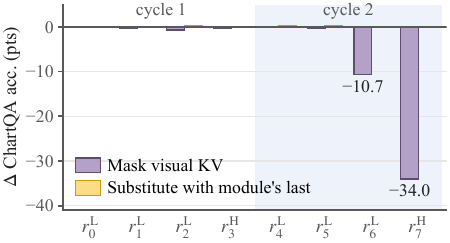}
\caption{\textbf{Invocation-level KV ablation} on ChartQA ($n{=}300$). Paired accuracy change relative to FullKV.}
\label{fig:slots}
\end{figure}

\paragraph{Invocation-level KV ablation.}
For each invocation $r$, we either mask its prefix visual KV during decoding or substitute it with the KV of the last invocation of the same module, keeping all other invocations unchanged (\cref{fig:slots}).
Masking the last invocations $r_6$ and $r_7$ costs 10.7 and 34.0 points, whereas masking any earlier one costs at most 0.7, and substituting it costs nothing.
Masking $r_0$--$r_5$ jointly costs 2.2 points (\cref{tab:stage}), so early invocations still need the image, just not their own copy.

\begin{table}[t]
\centering
\small
\setlength{\tabcolsep}{2pt}
\caption{\textbf{Where to share.} Accuracy change relative to FullKV of the same run (ChartQA full test set; DocVQA fixed 2000-sample subset, ANLS).}
\label{tab:stage}
\begin{adjustbox}{max width=\linewidth}\begin{tabular}{l l c c}
\toprule
\textbf{Policy} & \textbf{Stage} & \textbf{ChartQA} & \textbf{DocVQA} \\
\midrule
Share visual KV           & prefill  & $-$65.5 & $-$72.0 \\
Evict visual KV                           & decoding & $-$44.9 & $-$52.2 \\
Read the first invocation & decoding & $-$27.3 & $-$46.5 \\
\midrule
Skip visual KV in $r_0$--$r_5$ & decoding & $-$2.24 & $-$3.58 \\
\ours Read the last (\methodlite) & decoding & $-$0.08 & $-$0.18 \\
Skip visual KV in $r_0,r_1,r_4,r_5$ & decoding & $-$1.68 & $-$2.15 \\
\ours Read the last per cycle (\method) & decoding & $+$0.04 & $-$0.29 \\
\bottomrule
\end{tabular}\end{adjustbox}
\end{table}

\paragraph{Where to share.}
\Cref{tab:stage} varies when the shared copy is used and where it comes from.
Sharing during prefill is the most harmful, more so than reading early states at decoding ($-65.5$ vs.\ $-27.3$ on ChartQA), because it also breaks the iterative interaction between visual and question tokens.
Evicting visual KV removes needed evidence.
In the lower block, each pair stores the same visual copies. Skipping visual KV in the folded invocations costs 1.7--3.6 points, whereas reading the kept copy costs at most 0.3, so folding keeps visual information that the early invocations use.

\subsection{Ablation Studies}
\label{sec:ablation}

\begin{table*}[t]
\centering
\small
\caption{\textbf{Ablations.} Accuracy change relative to FullKV of the same run; defaults \colorbox{oursbg}{shaded}.
(a) \methodlite, full test sets; (b) ChartQA, 300 samples; (c) \methodlite, RTX 4090D.}
\label{tab:ablation}
\begin{minipage}[t]{0.31\linewidth}
\centering
\textbf{(a)} Shared tokens\\[2pt]
{\footnotesize\setlength{\tabcolsep}{3.5pt}\begin{adjustbox}{max width=\linewidth}\begin{tabular}{l c c c}
\toprule
\textbf{Tokens} & \textbf{Mem.} & \textbf{ChartQA} & \textbf{MathV.} \\
\midrule
\ours Visual   & 0.27 & $-$0.08 & $-$0.16 \\
Question                    & 0.98 & $-$0.88 & $-$0.86 \\
Visual + question           & 0.25 & $-$3.20 & $-$1.09 \\
\bottomrule
\end{tabular}\end{adjustbox}}
\end{minipage}\hfill
\begin{minipage}[t]{0.31\linewidth}
\centering
\textbf{(b)} Retained invocation\\[2pt]
{\footnotesize\setlength{\tabcolsep}{3.5pt}\begin{adjustbox}{max width=\linewidth}\begin{tabular}{l c c c}
\toprule
\textbf{Granularity} & \textbf{First} & \textbf{Avg.} & \textbf{Last} \\
\midrule
Model-wise (1)  & $-$43.67 & $-$39.67 & $-$25.33 \\
Module-wise (2) & $-$29.33 & $-$1.67 & 0.00 \\
\ours Cycle-wise (4)  & $-$1.00 & $+$0.67 & 0.00 \\
\bottomrule
\end{tabular}\end{adjustbox}}
\end{minipage}\hfill
\begin{minipage}[t]{0.31\linewidth}
\centering
\textbf{(c)} Keys, values and layers\\[2pt]
{\footnotesize\setlength{\tabcolsep}{3.5pt}\begin{adjustbox}{max width=\linewidth}\begin{tabular}{l c c c}
\toprule
\textbf{Shared} & \textbf{Mem.} & \textbf{ChartQA} & \textbf{DocVQA} \\
\midrule
FullKV (acc.)        & 1.00      & 74.52     & 78.44 \\
\midrule
Keys only            & 0.64 & $-$0.20 & $-$0.10 \\
Values only          & 0.64 & $-$0.16 & $+$0.03 \\
Shallow half layers  & 0.64 & $-$0.12 & $-$0.41 \\
Deep half layers     & 0.64 & $+$0.04 & $-$0.19 \\
\ours All (keys, values) & 0.27 & $-$0.20 & $-$0.64 \\
\bottomrule
\end{tabular}\end{adjustbox}}
\end{minipage}
\end{table*}
\begin{table}[t]
\centering
\footnotesize
\setlength{\tabcolsep}{4pt}
\caption{\textbf{Two design rules behind \method} (ChartQA, full test set).
\textbf{Cross.}: crosses modules. \emph{visual}: on visual KV only.}
\label{tab:rules}
\begin{adjustbox}{max width=\linewidth}\begin{tabular}{l c c r}
\toprule
\textbf{Mapping} & \textbf{Cross.} & \textbf{Mem.} & $\Delta$\textbf{Acc.} \\
\midrule
\multicolumn{4}{l}{\emph{Rule 1: fold within modules}} \\
Last-step sharing (Ouro), visual      & yes & 0.151 & $-$23.12 \\
Zero-shot sharing, $k{=}2$, visual    & yes & 0.272 & $-$1.12 \\
\methodlite (module-wise)             & no  & 0.272 & $-$0.08 \\
\midrule
\multicolumn{4}{l}{\emph{Rule 2: fold visual tokens only}} \\
Zero-shot sharing, $k{=}4$, whole prefix & no & 0.50  & $-$3.60 \\
Zero-shot sharing, $k{=}4$, visual       & no & 0.515 & $+$0.04 \\
\bottomrule
\end{tabular}\end{adjustbox}
\end{table}

\paragraph{Two design rules.}
\Cref{tab:rules} isolates the two rules behind \method with transferred baselines that violate them.
Crossing module boundaries costs accuracy even on visual KV alone. At the same memory as \methodlite, the Huginn $k{=}2$ mapping crosses module boundaries and loses 1.1 points, and last-step sharing, which serves every invocation from $r_7$, loses 23.1.
Sharing question KV costs accuracy as well. Huginn $k{=}4$ stays within modules but also shares question KV and loses 3.6 points, while on visual KV alone it shows no change.

\paragraph{Shared tokens.}
Folding question KV is costlier than folding visual KV on both ChartQA and MathVision, and folding both is worse still (\cref{tab:ablation}(a)). Visual KV is also where the memory is, so \method folds visual tokens only.

\paragraph{Retained invocation.}
Keeping the last invocation is lossless at both granularities.
Keeping the first costs 29.3 points at module-wise granularity but only 1.0 at cycle-wise granularity (\cref{tab:ablation}(b)), where the kept and the folded invocations are closer in depth. This makes the cycle-wise granularity the safer default.
Folding across modules fails whichever invocation is kept.

\paragraph{Keys, values and layers.}
Folding only keys, only values or half of the layers (\cref{tab:ablation}(c)) stores 64\% of the prefix KV and stays slightly closer to FullKV on DocVQA. Full folding stores $2.4\times$ less at a cost of about half a point there, so we fold all of them.

\subsection{Efficiency}
\label{sec:efficiency}

\begin{figure}[t]
\centering
\includegraphics[width=\linewidth]{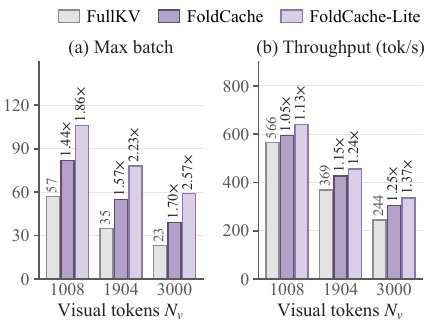}
\caption{\textbf{Decoding efficiency} on one A800 (BF16, 512 output tokens, preallocated cache).
(a) Largest batch fitting in the same memory; (b) throughput at that batch. Labels give FullKV values and ratios to FullKV.}
\label{fig:efficiency}
\end{figure}

Caching removes the recomputation of cache-free decoding. With 1{,}904 visual tokens, the time per output token drops from 441 to about 85\,ms at 2{,}048 output tokens.
Folding frees the memory that a full cache spends on repeated visual states.
In the same memory as FullKV, \method fits 1.4--1.7$\times$ larger batches and reaches 1.05--1.25$\times$ its throughput, and \methodlite fits 1.9--2.6$\times$ larger batches at 1.13--1.37$\times$ the throughput (\cref{fig:efficiency}).
The gains grow with the number of visual tokens, whose KV dominates the cache, and \methodlite fits as many images with 3{,}000 visual tokens (batch 59) as FullKV with 1{,}008 (batch 57).
Single-request latency is in the supplementary material.

\subsection{Text Tokens in a Looped Language Model}
\label{sec:ouro}

Ouro-1.4B ($R{=}4$), a looped language model without visual tokens, shows what happens when text KV is shared across recurrent steps.
At decoding, we replace the prompt KV of each step with that of the last step and keep the answer KV.
GSM8K accuracy drops from 78.8 to 6.6, and replacing the prompt KV of the first step alone costs 72 points, whereas replacing that of the second or third step costs 2.6 or 1.4 points (supplementary material). Replacing the prompt KV of the first step is even worse than dropping it, which costs 54.6 points.
The cost falls with the distance to the last step, in line with the first-order view of \cref{sec:observations}.
This matches the cost of folding question KV on LoopVL (\cref{tab:ablation}(a)) and supports keeping question and answer KV exact.

\section{Conclusion}
\label{sec:conclusion}

Recurrence gives vision-language models depth without parameters, but it multiplies the KV cache by the number of invocations, and in VLMs most of this cache stores the image.
We showed that this recurrent axis is redundant at decoding time. After an exact prefill, the visual KV of a module's last invocation can stand in for that of its earlier invocations.
Earlier invocations still use the image. Skipping their visual KV costs 1.7--3.6 points, and reading the kept copy matches FullKV.
A first-order view explains this rule. Later computation depends most on the last invocations, and the error of reading another invocation's copy grows with the distance to it, which is largest across modules.
\method turns this observation into a training-free cache that halves the prefix KV at FullKV accuracy, composes with quantization and token eviction, and fits up to $1.7\times$ larger batches in the same memory ($2.6\times$ for \methodlite).
We hope these findings help future looped models keep fewer states and spend the saved memory on deeper recurrence or higher image resolution.

{
    \small
    \bibliographystyle{ieeenat_fullname}
    \bibliography{references}
}

\clearpage
\appendix
\twocolumn[{\centering\Large\bfseries Supplementary Material\par\vspace{1em}}]
\setcounter{table}{0}
\setcounter{figure}{0}
\setcounter{equation}{0}
\renewcommand{\theHtable}{supp.\arabic{table}}
\renewcommand{\theHfigure}{supp.\arabic{figure}}
\renewcommand{\theHequation}{supp.\arabic{equation}}
\renewcommand{\thetable}{S\arabic{table}}
\renewcommand{\thefigure}{S\arabic{figure}}
\renewcommand{\theequation}{S\arabic{equation}}
\setcounter{dbltopnumber}{3}

\section{Transferred Baselines That Also Share Answer KV}
\label{sec:suppl_scope}

In the main paper, all transferred baselines share prefix KV only and keep answer-token KV exact, matching \method.
\Cref{tab:suppl_scope} reports the variants that, as in their original form, also share the KV of generated answer tokens across invocations.
Sharing answer KV consistently increases the loss, so the prefix-only variants in \cref{tab:main} are the stronger baselines.

\begin{table}[h]
\centering
\footnotesize
\caption{\textbf{Sharing scope of transferred baselines} on ChartQA (full test set, $n{=}2500$). Accuracy change relative to FullKV of the same run.}
\label{tab:suppl_scope}
\begin{adjustbox}{max width=\linewidth}\begin{tabular}{l c c}
\toprule
\textbf{Method} & \textbf{Prefix only} & \textbf{Prefix + answer} \\
\midrule
Last-step sharing (Ouro)              & $-$26.32 & $-$34.40 \\
Zero-shot sharing, $k{=}2$ (Huginn) & $-$4.48  & $-$6.72 \\
Zero-shot sharing, $k{=}4$ (Huginn) & $-$3.60  & $-$4.56 \\
\bottomrule
\end{tabular}\end{adjustbox}
\end{table}

\section{Multiple-Choice Benchmarks}
\label{sec:suppl_mc}

KV compression applied after prefill cannot change an answer determined by the first generated token, because that token is produced by the uncompressed prefill.
\Cref{tab:suppl_mc} confirms this on four multiple-choice benchmarks: both \method and visual KV eviction leave every benchmark unchanged, even though eviction degrades ChartQA by 44.9 points.
These benchmarks therefore serve only as a sanity check.

\begin{table}[h]
\centering
\footnotesize
\caption{\textbf{Multiple-choice benchmarks.} Accuracy of FullKV and paired change of each policy.}
\label{tab:suppl_mc}
\begin{adjustbox}{max width=\linewidth}\begin{tabular}{l c c c c}
\toprule
\textbf{Method} & \textbf{MMStar} & \textbf{AI2D} & \textbf{VMCBench} & \textbf{RealWorldQA} \\
\midrule
FullKV              & 62.53 & 75.68 & 71.00 & 71.24 \\
Visual KV eviction  & 0.00  & 0.00  & 0.00  & 0.00 \\
\methodlite & 0.00 & 0.00 & 0.00 & 0.00 \\
\bottomrule
\end{tabular}\end{adjustbox}
\end{table}

\section{Equivalence Checks}
\label{sec:suppl_equiv}

\paragraph{KV cache versus recomputation.}
LoopVL decodes without a KV cache and recomputes the full recurrence for every token.
Our FullKV implementation caches the KV states of all $R{=}8$ invocations.
On 50 samples, its generated tokens match those of the original cache-free decoding in all 50 cases.

\paragraph{Segmented attention.}
\method computes attention separately over the shared visual segment and the per-invocation segments, and merges them by their log-sum-exp.
At batch size 1, on 50 real samples, the generated tokens match those of standard attention over the concatenated cache in all 50 cases, token by token. The largest logit difference is at most 0.22, because the accumulation order differs.
At batch size 4, with 4 synthetic inputs and 16 forced decoding steps, the rows that match their batch-size-1 outputs are 2/4 for FullKV computed with the segmented kernel (FullKV-Seg), 3/4 for \method and 3/4 for \methodlite.
Uncompressed FullKV-Seg diverges as well, so these differences come from the numerics of the batched kernel rather than from folding. For the same reason, long-form generation is compared with FullKV-Seg (\cref{sec:suppl_long}).
The efficiency results in \cref{sec:efficiency} use this segmented implementation, which reads the shared visual segment in place (\cref{sec:suppl_latency}).

\section{Earlier Prefill-Stage Controls}
\label{sec:suppl_hist}

An earlier version of our implementation evaluated several controls that intervene during prefill, and \cref{tab:stage} reports the prefill-sharing control re-run with the current code.
\Cref{tab:suppl_hist} lists the earlier results for reference, as five-benchmark averages over MMStar, RealWorldQA, VMCBench, AI2D and ChartQA, rescored with the current evaluation code.
All of them lose 15--39 points, consistent with the finding that prefill must stay exact.

\begin{table}[h]
\centering
\small
\caption{\textbf{Earlier prefill-stage controls} (earlier run, rescored with the current evaluation code). Five-benchmark average accuracy.}
\label{tab:suppl_hist}
\begin{tabular}{l c}
\toprule
\textbf{Prefill policy} & \textbf{Avg.} \\
\midrule
No intervention (FullKV)                                   & 71.18 \\
Block visual updates in later $f_L$ invocations            & 56.00 \\
Freeze visual states after cycle 1                         & 49.99 \\
Read first-invocation visual states (fine-tuned)           & 52.75 \\
Share one visual state across invocations                  & 32.01 \\
\bottomrule
\end{tabular}
\end{table}

\section{Full-Set Results}
\label{sec:suppl_fullset}

\Cref{tab:main} reports DocVQA, InfoVQA and TextVQA on fixed 2000-sample subsets, for which all baselines were evaluated.
On the full validation sets, \method stays within 0.1 points of FullKV of the same run, with changes of $-0.05$ on DocVQA (5349 samples), $+0.09$ on TextVQA (5000 samples) and $-0.04$ on InfoVQA (2801 samples). \methodlite reduces accuracy by 0.44 points on DocVQA and 1.18 points on TextVQA. On TextVQA answers of at least two tokens, the loss is 1.35 points, and this subset does not pass the non-inferiority test defined in \cref{sec:setup}.
Combined with INT4 quantization, \methodlite stores the least prefix KV of all configurations (about $0.09\times$), but on TextVQA it is 1.13 points below KIVI-INT2 (\cref{tab:suppl_int2}).

\section{MiniCache-Style Averaging without SLERP}
\label{sec:suppl_minicache}

The main paper adapts MiniCache with SLERP interpolation and token retention.
\Cref{tab:suppl_minicache} compares it with plain averaging of adjacent invocations within a module, which keeps no tokens unmerged.

\begin{table*}[t]
\centering
\small
\setlength{\tabcolsep}{5pt}
\caption{\textbf{MiniCache-style merging of adjacent invocations} (A800, same protocol as \cref{tab:main}). Accuracy and difference to FullKV of the same run.}
\label{tab:suppl_minicache}
\begin{tabular}{l c c c c c c c}
\toprule
\textbf{Variant} & \textbf{Mem.} & \textbf{ChartQA} & \textbf{MathV.} & \textbf{DocVQA} & \textbf{InfoVQA} & \textbf{TextVQA} & \textbf{Avg.} \\
\midrule
Averaging    & 0.52 & 73.56\dd{$-$1.16} & 38.12\dd{$-$0.13} & 71.30\dd{$-$7.14} & 57.26\dd{$-$3.02} & 51.79\dd{$-$1.02} & 58.41\dd{$-$2.50} \\
SLERP + 5\% retention & 0.55 & 73.96\dd{$-$0.76} & 38.22\dd{$-$0.03} & 72.16\dd{$-$6.28} & 57.38\dd{$-$2.89} & 51.80\dd{$-$1.00} & 58.70\dd{$-$2.20} \\
\bottomrule
\end{tabular}
\end{table*}

\section{Folding plus INT4 versus INT2 Quantization}
\label{sec:suppl_int2}

\Cref{tab:suppl_int2} lists all direct paired comparisons between folding plus INT4 and INT2 quantization in the configuration of KIVI (keys per channel and values per token, group size 32, the last 128 prefix tokens kept in BF16), on the same samples and in the same run as \cref{tab:main}.

\begin{table*}[t]
\centering
\small
\setlength{\tabcolsep}{4pt}
\caption{\textbf{Direct paired comparisons with INT2 quantization} (same run as \cref{tab:main}).
Differences in points (first minus second). Bold marks statistically significant differences under the paired-bootstrap protocol in \cref{sec:setup}.
\textbf{Mem.}: prefix KV relative to FullKV, including quantization parameters and the BF16 window.}
\label{tab:suppl_int2}
\begin{adjustbox}{max width=\textwidth}
\begin{tabular}{l c c c c c}
\toprule
\textbf{Comparison} & \textbf{ChartQA} & \textbf{MathVision} & \textbf{DocVQA} & \textbf{InfoVQA} & \textbf{TextVQA} \\
\midrule
\method{}\,+\,INT4 $-$ KIVI-INT2           & $+$0.36 & $-$0.07 & $-$0.03 & $+$0.14 & $-$0.32 \\
\method{}\,+\,INT4 $-$ FlashLoop-INT2      & $+$0.32 & $-$0.16 & $+$0.25 & $\mathbf{+0.76}$ & $+$0.17 \\
\methodlite{}\,+\,INT4 $-$ KIVI-INT2       & $+$0.20 & $-$0.20 & $+$0.09 & $-$0.30 & $\mathbf{-1.13}$ \\
\methodlite{}\,+\,INT4 $-$ FlashLoop-INT2  & $+$0.16 & $-$0.30 & $+$0.36 & $+$0.33 & $-$0.65 \\
\midrule
\multicolumn{6}{l}{\emph{Mem.}} \\
KIVI-INT2 / FlashLoop-INT2               & 0.247 & 0.296 & 0.238 & 0.239 & 0.238 \\
\method{}\,+\,INT4                         & 0.167 & 0.266 & 0.161 & 0.162 & 0.159 \\
\methodlite{}\,+\,INT4                     & 0.098 & 0.205 & 0.092 & 0.093 & 0.091 \\
\bottomrule
\end{tabular}\end{adjustbox}
\end{table*}

\paragraph{Sensitivity to the INT2 group size.}
With a group size of 64 instead of 32, INT2 quantization stores less ($0.16\times$) but loses more accuracy (\cref{tab:suppl_int2g64}).

\begin{table*}[t]
\centering
\small
\setlength{\tabcolsep}{5pt}
\caption{\textbf{INT2 quantization with group size 64} (same protocol as \cref{tab:main}). Accuracy; FullKV of \cref{tab:main} is listed for reference.}
\label{tab:suppl_int2g64}
\begin{tabular}{l c c c c c c c}
\toprule
\textbf{Method} & \textbf{Mem.} & \textbf{ChartQA} & \textbf{MathV.} & \textbf{DocVQA} & \textbf{InfoVQA} & \textbf{TextVQA} & \textbf{Avg.} \\
\midrule
FullKV & 1.00 & 74.72 & 38.26 & 78.44 & 60.28 & 52.81 & 60.90 \\
\midrule
KIVI-INT2 (g64)      & 0.16 & 73.32 & 37.37 & 76.94 & 59.12 & 51.60 & 59.67 \\
FlashLoop-INT2 (g64) & 0.16 & 74.00 & 37.86 & 77.34 & 59.27 & 51.63 & 60.02 \\
\bottomrule
\end{tabular}
\end{table*}

\section{Sharing Scopes on Ouro-1.4B}
\label{sec:suppl_ouro}

Ouro-1.4B ($R{=}4$) has no visual tokens, so these experiments test only whether \emph{text} KV can be shared across recurrent steps.
All runs use GSM8K test ($n{=}1319$), 3-shot prompting, greedy decoding and at most 256 generated tokens. We report flexible matching as the main metric and strict matching for reference.
Differences are relative to FullKV of the same run.
Interventions are applied only at decoding, and prefill is unchanged.

\paragraph{Sharing scope.}
\Cref{tab:ouro} replaces the step-specific KV of every step with that of the last step $r_3$, for the prompt tokens only (last-prefix), for the generated tokens only (last-gen), or for both (last-all).
Every scope loses accuracy, and sharing the prompt KV costs the most.
Under teacher forcing, last-prefix changes only 12\% of the next-token predictions, yet in free generation these errors accumulate to a loss of 72 points.
Strict accuracy rescores the existing generations without regenerating, taking the first match of \texttt{The answer is \textbackslash\$?(-?[0-9.,]+)}. With this parser, our FullKV reaches 78.17, close to the 78.92 reported by Ouro.
Each configuration is paired per problem with FullKV of the same run, and the FullKV outputs of the two runs match on every problem.
Ouro does not fully specify which KV states are shared during decoding, and its description is closest to last-all.

\begin{table*}[t]
\centering
\small
\setlength{\tabcolsep}{6pt}
\caption{\textbf{Sharing scope of last-step KV} in Ouro-1.4B at decoding (GSM8K, $n{=}1319$).
Accuracy and change relative to FullKV.
\textbf{Token agr.}: teacher-forced next-token agreement with FullKV; \textbf{KL}: mean per-token KL divergence to FullKV.}
\label{tab:ouro}
\begin{tabular}{l c c c c}
\toprule
\textbf{Shared KV} & \textbf{Flexible} & \textbf{Strict} & \textbf{Token agr.} & \textbf{KL} \\
\midrule
None (FullKV)                & 78.77 & 78.17 & 100 & 0 \\
Prompt (last-prefix)         & 6.60\dd{$-$72.18} & 4.93\dd{$-$73.24} & 87.9 & 0.385 \\
Generated tokens (last-gen)  & 68.92\dd{$-$9.86} & 68.92\dd{$-$9.25} & 96.8 & 0.078 \\
Both (last-all)              & 0.15\dd{$-$78.62} & 0.00\dd{$-$78.17} & 44.0 & 2.46 \\
\bottomrule
\end{tabular}
\end{table*}

\paragraph{Per-step interventions.}
\Cref{tab:suppl_ouro13} intervenes on the prompt KV of individual steps, and the KV of generated tokens is never changed.
\emph{Replace $r_k$} substitutes the prompt KV of step $r_k$ with that of the last step $r_3$, and \emph{Drop $r_k$} removes it. Interventions on several steps are applied jointly.
Two identity checks pass on all 1319 problems: replacing $r_3$ by itself reproduces the token sequences and texts of FullKV, and replacing $r_0$--$r_2$ those of last-prefix.
Replacing the prompt KV of $r_0$ alone costs about as much as replacing all early steps, whereas replacing $r_1$ or $r_2$ costs 1--3 points.
Unlike the visual KV of LoopVL, the first step's text KV in Ouro cannot be substituted by the last step's.

\begin{table*}[t]
\centering
\small
\setlength{\tabcolsep}{8pt}
\caption{\textbf{Per-step interventions on prompt KV} in Ouro-1.4B at decoding (GSM8K, $n{=}1319$; same protocol as \cref{tab:ouro}).
Only the prompt KV is modified, and strict accuracy is rescored as in \cref{tab:ouro}.}
\label{tab:suppl_ouro13}
\begin{tabular}{l c c}
\toprule
\textbf{Configuration} & \textbf{Flexible} & \textbf{Strict} \\
\midrule
FullKV      & 78.77 & 78.17 \\
\midrule
Replace $r_0$ & 6.67\dd{$-$72.10}  & 5.38\dd{$-$72.78} \\
Replace $r_1$ & 76.19\dd{$-$2.58}    & 75.13\dd{$-$3.03} \\
Replace $r_2$ & 77.41\dd{$-$1.36}    & 76.42\dd{$-$1.74} \\
Replace $r_3$ (identity) & 78.77\dd{0.00}                     & 78.17\dd{0.00} \\
Replace $r_0$--$r_2$ & 6.60\dd{$-$72.18}  & 4.93\dd{$-$73.24} \\
Last-prefix (identical to previous row) & 6.60\dd{$-$72.18}  & 4.93\dd{$-$73.24} \\
\midrule
Drop $r_0$ & 24.18\dd{$-$54.59} & 22.29\dd{$-$55.88} \\
Drop $r_1$ & 52.24\dd{$-$26.54} & 50.87\dd{$-$27.29} \\
Drop $r_2$ & 46.63\dd{$-$32.15} & 44.28\dd{$-$33.89} \\
Drop $r_3$ & 6.90\dd{$-$71.87}  & 5.61\dd{$-$72.55} \\
Drop $r_0$--$r_2$ & 0.08\dd{$-$78.70}  & 0.00\dd{$-$78.17} \\
Drop $r_0$--$r_3$ & 0.45\dd{$-$78.32}  & 0.00\dd{$-$78.17} \\
\bottomrule
\end{tabular}
\end{table*}

\section{Long-Form Generation}
\label{sec:suppl_long}

\begin{table}[!htb]
\centering
\footnotesize
\setlength{\tabcolsep}{3.5pt}
\caption{\textbf{Long-form generation} on ChartQA-CoT ($n{=}2500$, up to 512 tokens).
\boldmath$\Delta$\unboldmath: accuracy difference to FullKV-Seg.
\textbf{Trunc.}: answers hitting the 512-token limit; \textbf{Agree.}: final answer identical to FullKV-Seg.
The first row (SDPA kernel) measures numerical noise.}
\label{tab:longform}
\begin{adjustbox}{max width=\linewidth}\begin{tabular}{l c c c c}
\toprule
\textbf{Method} & \textbf{Acc.} & \boldmath$\Delta$ \textbf{vs.\ FullKV-Seg} & \textbf{Trunc.} & \textbf{Agree.} \\
\midrule
FullKV (SDPA kernel)          & 50.40 & $-$0.32 & 121 & 85.8 \\
FullKV-Seg (reference)       & 50.72 & --                         & 129 & 100.0 \\
\midrule
\ours \textbf{\method (default)} & 53.24 & $+$2.52 & 116 & 66.1 \\
\methodlite                   & 50.84 & $+$0.12 & 117 & 59.8 \\
\bottomrule
\end{tabular}\end{adjustbox}
\end{table}

With LoopVL's official reasoning trigger, ChartQA answers become chains of thought of up to 512 generated tokens, so most of each answer is generated from the folded cache.
We compare every configuration with FullKV computed by the same segmented kernel (FullKV-Seg) and use the paired difference in relaxed accuracy as the main metric, with the non-inferiority margin of $-1.5$ points used in the main paper. Decoding is greedy.
Truncated answers are scored on their last line (one answer of \method cannot be parsed, none of the others).

\begin{figure}[t]
\centering
\includegraphics[width=\linewidth]{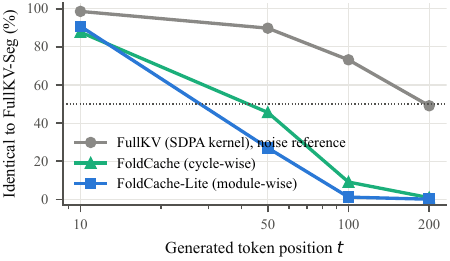}
\caption{\textbf{Token-level agreement} with FullKV-Seg during chain-of-thought generation (ChartQA, $n{=}2500$): fraction of trajectories identical to FullKV-Seg up to generated position $t$.
The gray curve (FullKV with the SDPA kernel) is the numerical-noise reference.}
\label{fig:survival}
\end{figure}

Agreement is not a suitable main metric for long generation.
A trajectory diverges after the first changed token, so any perturbation, however small, is carried to the end of the answer.
Changing only the attention kernel, from SDPA to the segmented kernel, already lets about half of the trajectories diverge by about 200 generated tokens (gray curve in \cref{fig:survival}), although accuracy changes by only 0.32 points (\cref{tab:longform}).
Token-level agreement thus reflects sensitivity to any perturbation rather than information loss. We report answer agreement and the survival curve only as auxiliary measures of how early trajectories diverge, and judge long-form accuracy by the paired difference.

\method passes the non-inferiority test against FullKV-Seg, with an accuracy difference of $+2.52$ points. Excluding samples truncated by either configuration leaves a difference of $+2.74$ points, indicating that truncation does not explain the result. We use this result to establish non-inferiority, without claiming an accuracy improvement.
\methodlite has a small positive difference ($+0.12$ points) but does not pass the same test.
\method also passes the test on the human-written and augmented subsets, with differences of $+0.88$ and $+4.16$ points, respectively. \methodlite loses 1.60 points on the harder human-written questions, consistent with its sensitivity on text-dense images.

\section{Concatenation versus Segmented Attention}
\label{sec:suppl_latency}

\paragraph{Log-sum-exp merge.}
For a decoding query $q$ and a segment $s$ with keys $K_s$ and values $V_s$, the attention kernel returns $o_s=\mathrm{softmax}(\alpha qK_s^\top)V_s$ and the log-sum-exp $\ell_s=\log\sum_j\exp(\alpha qk_{s,j}^\top)$, with attention scale $\alpha$.
Since $e^{\ell_s}$ is the softmax normalizer of segment $s$, attention over the union of the visual segment $v$ and the text segment $t$ is
\begin{equation}
o=\frac{e^{\ell_v-m}o_v+e^{\ell_t-m}o_t}{e^{\ell_v-m}+e^{\ell_t-m}},\qquad m=\max(\ell_v,\ell_t).
\end{equation}
We compute $o_s$ and $\ell_s$ with the flash attention kernel of PyTorch, merge them in FP32 after subtracting $m$, and cast the result to BF16 before the output gate and projection of LoopVL.
Keys already carry rotary position embeddings, so the segment order does not matter.
Each group keeps its visual segment in a preallocated $[B,H,N_v,D]$ buffer, and each invocation writes generated tokens in place into its own text buffer, after its header and question KV.

\paragraph{Concatenation.}
A simpler implementation copies the shared visual segment next to the per-invocation question and answer KV at every decoding step and runs standard attention over the concatenation.
It generates the same tokens, but every step materializes a full-size copy of the visual segment for each invocation and batch element, so the shared copy no longer saves memory traffic, and the transient copies, not the stored cache, bound the batch.
In our measurements this implementation reaches only about $0.3\times$ the throughput of the segmented one.
We therefore use the segmented implementation throughout: it reads the shared visual base in place, and merging two attention segments by their log-sum-exp is exact up to floating-point accumulation order (\cref{sec:suppl_equiv}).

\paragraph{Single-request latency.}
At batch size 1 (A800, 512 output tokens), the time per output token is about 85\,ms for FullKV and about 103\,ms for both \method and \methodlite.
The difference is a fixed per-step cost: each invocation runs two attention segments and merges them by their log-sum-exp instead of a single attention call.
At batch size 1 the decoding step is not memory-bound, so this overhead is not offset by the smaller cache. At larger batches, the memory saved by folding dominates (\cref{fig:efficiency}).
A fused kernel that computes both segments in one pass would remove this overhead.

\paragraph{Maximum batch.}
Prefill runs in chunks of 4 samples. The batch in \cref{fig:efficiency} counts the KV of all generated tokens and, with folding, is limited by prefill activations.

\section{Evaluation Details}
\label{sec:suppl_eval}

\paragraph{Datasets.}
DocVQA, InfoVQA and TextVQA use the validation splits released by lmms-lab, with the same sample counts as the official splits (5349, 2801 and 5000).
Prompts and metrics follow VLMEvalKit.

\paragraph{Long-form generation.}
We first evaluated MathVision with generation forced to 512 tokens.
This setting turned out to be unusable: most samples end within a few tokens, so most positions are continuations after the end-of-sequence token, and the remaining chains of thought often degenerate into repetition.
Token-level agreement on this setting cannot distinguish \method from visual KV eviction.
We therefore use ChartQA with LoopVL's official reasoning trigger (\cref{sec:suppl_long}), where samples produce natural chains of thought of up to 512 generated tokens.

\paragraph{Run-to-run variation.}
All comparisons in the paper are paired within the same run.
Across runs, accuracy varies slightly due to numerical differences. For example, FullKV obtains 62.53 on MMStar in the current run and 63.40 in an earlier run.
The main paper reports the current run throughout.

\section{Limitations and Future Work}
\label{sec:suppl_limits}

Validating the rule on other hierarchical looped VLMs, once their weights are public, is left for future work.
The aggressive granularity is more sensitive on text-dense images (\cref{sec:suppl_fullset}).
The reliance of answer tokens on the last states also suggests self-speculative decoding with shallow drafts and full-depth verification.

\end{document}